# DeepSperm: A robust and real-time bull sperm-cell detection in densely populated semen videos

Priyanto Hidayatullah[1*], Xueting Wang[2], Toshihiko Yamasaki[2], Tati L.E.R. Mengko[1], Rinaldi Munir[1], Anggraini Barlian[3], Eros Sukmawati[4], Supraptono Supraptono[4]

[1]School of Electrical Engineering and Informatics, Institut Teknologi Bandung,
[2]Department of Information and Communication Engineering, The University of Tokyo,
[3]School of Life Sciences and Technology, Institut Teknologi Bandung
[4]Balai Inseminasi Buatan Lembang, Bandung

priyanto@polban.ac.id, xt_wang@hal.t.u-tokyo.ac.jp, yamasaki@hal.t.u-tokyo.ac.jp, tati@stei.itb.ac.id, rinaldi@informatika.org, aang@sith.itb.ac.id, ros.sukmawati@gmail.com, supraptono1210@gmail.com

## Abstract

**Background and Objective:** Object detection is a primary research interest in computer vision. Sperm-cell detection in a densely populated bull semen microscopic observation video presents challenges such as partial occlusion, vast number of objects in a single video frame, tiny size of the object, artifacts, low contrast, and blurry objects because of the rapid movement of the sperm cells. This study proposes an architecture, called DeepSperm, that solves the aforementioned challenges and is more accurate and faster than state-of-the-art architectures.

**Methods:** In the proposed architecture, we use only one detection layer, which is specific for small object detection. For handling overfitting and increasing accuracy, we set a higher network resolution, use a dropout layer, and perform data augmentation on hue, saturation, and exposure. Several hyper-parameters are tuned to achieve better performance. We compare our proposed method with those of a conventional image processing-based object-detection method, you only look once (YOLOv3), and mask region-based convolutional neural network (Mask R-CNN).

**Results:** In our experiment, we achieve 86.91 mAP on the test dataset and a processing speed of 50.3 fps. In comparison with YOLOv3, we achieve an increase of 16.66 mAP point, 3.26 x faster on testing, and 1.4 x faster on training with a small training dataset, which contains 40 video frames. The weights file size was also reduced significantly, with 16.94 x smaller than that of YOLOv3. Moreover, it requires 1.3 x less graphical processing unit (GPU) memory than YOLOv3.

**Conclusions:** This study proposes DeepSperm, which is a simple, effective, and efficient architecture with its hyper-parameters and configuration to detect bull sperm cells robustly in real time. In our experiment, we surpass the state of the art in terms of accuracy, speed, and resource needs.

# 1   Introduction

Automatic sperm evaluation has been a critical problem. The intra-variabilities and inter-variabilities [1], its subjectivity [2], high time and human-resource consumptions, and exhaustion to the observer's eyes have been the main drawbacks of manual evaluation; a computer-aided sperm analysis (CASA) robust sperm detection capability is urgently required.

Several studies have been conducted on sperm-cell detection. However, some problems remain unsolved. In this study, we focus on solving these three problems: (1) limited accuracy, (2) high computational cost, and (3) limited annotated training data detailed in the following section.

The previous methods mostly used conventional image-processing approaches [3]–[5]. Some of the researchers employed mainly image binary morphological operations [6]–[8]. In contrast, Nissen et al. [9] used some sets of CNN architectures for performing sperm detection. In those studies, they claimed to have achieved an 84–94% sperm detection on low concentrated semen. However, this percentage significantly degrades on highly concentrated semen.

To summarize, the limited accuracy was caused by these specific challenges of sperm-cell detection on densely populated semen: frequent partial occlusion, vast number of objects in a single video frame, tiny size of the object, artifacts, low contrast, and blurry object because of the rapid movement of the sperm cell. These were the first problems we addressed.

Sperm-cell detection is a kind of object detection. Deep learning-based approaches, such as Mask R-CNN [10] and YOLOv3 [11], have achieved state-of-the-art performance for solving general object-detection problems. Unfortunately, owing to the fairly large size of the architecture, large weights files were produced that needed considerable amounts of training time, graphical processing unit (GPU) memory, and a considerable amount of storage space. In summary, the high computational cost was the second problem that we addressed.

Another problem with the deep learning-based object-detection methods is the need for a large amount of annotated training data to prevent overfitting and achieve adequate accuracy. Unfortunately, performing annotation in this study case was fairly laborious. In a single video frame, there can be as many as 500 cells. The limited annotated training data was the third problem that we addressed.

## 2  Materials and Methods

### 2.1  Dataset

There were six bull-sperm observation videos from six different bulls at the Balai Inseminasi Buatan Lembang, Indonesia. The samples were not stained. The length of the videos varied from 15 s to 124 s. In each video frame, the average number of sperms was

446.4. To capture the movement of the sperms, a phase-contrast microscope, with a total magnification of 100 x, was employed. The frame video resolution was 640 × 480 pixels recorded at 25 fps. Each video was taken in moderately different lighting conditions.

We built the training dataset from the first bull video sample, extracted the first 50 frames, shuffled the dataset randomly; 80% was used for the training dataset and the remainder for the validation dataset. We randomly extracted two video frames from each of the remaining videos and classified them as the test dataset. Table 1 illustrates the dataset proportions.

Table 1. Dataset proportions

| Samples | Video 1 | | Video 2 | Video 3 | Video 4 | Video 5 | Video 6 |
|---|---|---|---|---|---|---|---|
| | Training | Validation | Testing | | | | |
| Number of video frames | 40 | 10 | 2 | 2 | 2 | 2 | 2 |

To know the characteristics of the dataset, we converted the frame extracted from each video sample into a greyscale image and plotted its histogram as shown in Fig. 1. We also calculated the pixel mean value and the standard deviation ($std$) of the image pixels. There were two conclusions drawn from the samples; firstly, all the histograms were very narrow, yielding small standard deviation values, represent that all the samples had low contrast; secondly, the histogram of test dataset samples were more left-skewed than the histogram for the training/validation samples, yielding test dataset samples mean pixel values were lower than that of the training/validation dataset, represent that test dataset samples had lower luminance values than the training/validation dataset samples.

Fig. 1. (a) Training/validation dataset, mean = 171.19 and std = 27.92; (b) Testing video 1, mean = 158.99 and std = 25.99; (c) Testing video 2, mean = 148.86 and std = 22.8; (d) Testing video 3, mean = 88.56 and std = 21.17; (e) Testing video 4, mean = 93.09 and std = 22.50; (f) Testing video 5, mean = 102.86 and std = 21.46.

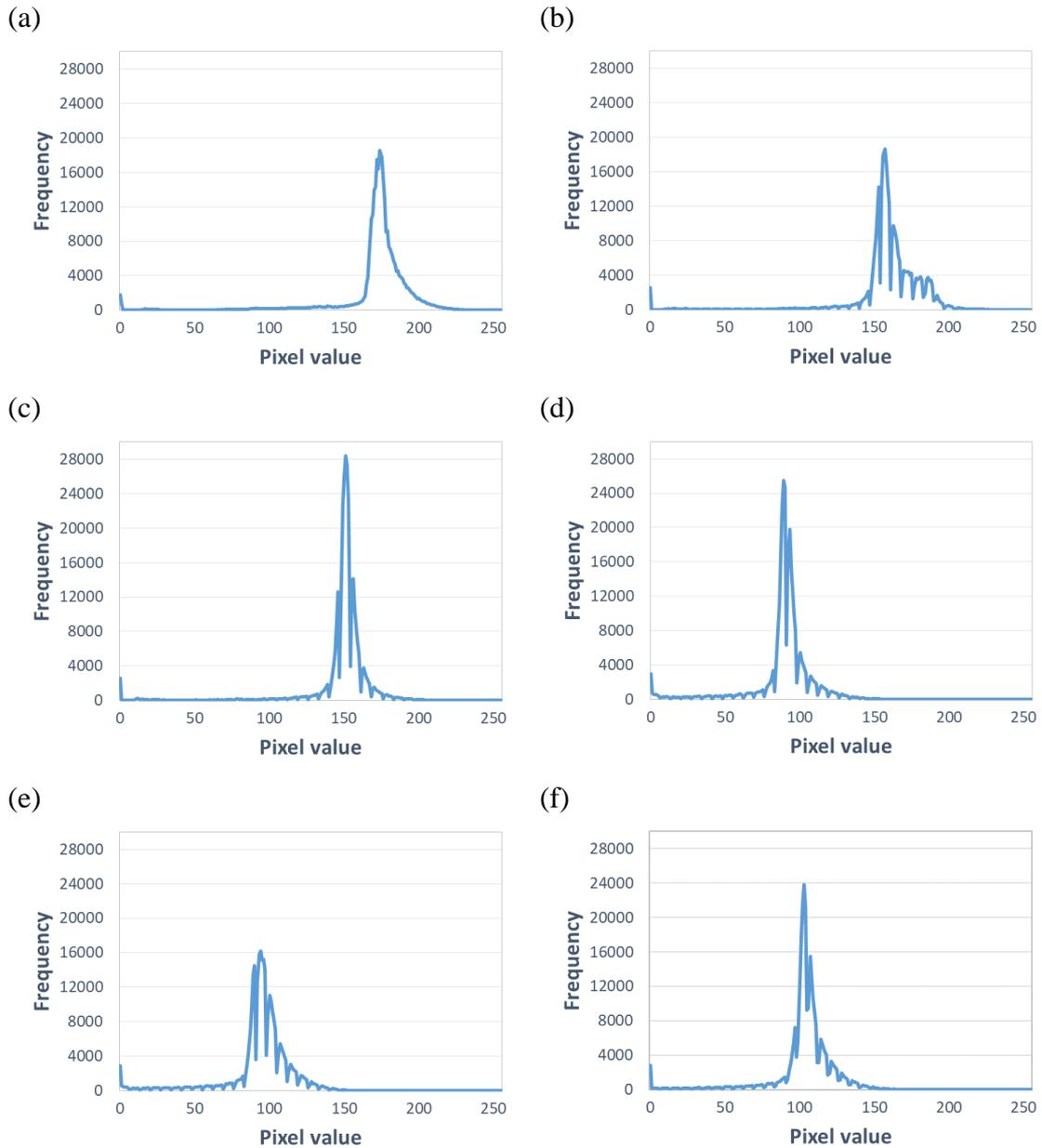

## 2.2 Ground truth and dataset annotation

Ground truth data were obtained from manual detection of sperms by two experienced veterinarians who have more than 13 years of experience. The dataset in this study was relatively small because its fairly laborious to annotate hundreds of sperm cells in a video frame. Several days were required to annotate 60 video frames. The tool used in [12] was adopted to annotate the dataset manually.

As an additional note, we evaluated sperms that reached the frame border. If 50% of their heads were visible, they were marked and counted in. In total, there were 18,882 sperm cells in the training dataset, 4,728 in the validation dataset, and 3,174 in the test dataset. Therefore, we had a considerable number of annotated objects, though there were only six videos in the dataset.

## 2.3 Architecture

The proposed architecture was based on YOLO. To improve the detection accuracy, the network resolution was increased up to $640 \times 640$. The network contained 42 layers in total. All the convolutional layers used batch normalization. The first seven layers were designed to downsample the image until a sufficient resolution for the architecture to detect small objects accurately ($80 \times 80$).

In the following layers, there were 24 deep convolutional layers with leaky RELU [13] activation function which formula presented in equation (1) [13], i.e.,

$$\emptyset(x) = \begin{cases} x, & if\ x > 0 \\ 0.1\ x, & otherwise. \end{cases} \tag{1}$$

To prevent vanishing/exploding gradient problems and to increase the detection accuracy, a shortcut layer was added to utilize the residual layers for every two convolutional layers [14]. The last layer, YOLO layer, gave detection prediction on each anchor box along with its confidence score. The number of filters of the YOLO layer was set according to equation (2) [13]. We used three anchor boxes, $1 \times 1$ grid size, and one class (sperm class). Therefore, the number of the filters at the YOLO layer according to equation (2) was 18.

$$n = S \times S \times (B * 5 + C), \tag{2}$$

where
$n$ = number of filters
$S \times S$ = number of the grids
$B$ = number of anchor boxes
$C$ = number of classes

At the end of the network, the logistic function was used. While YOLOv3 had three YOLO layers, there was only one YOLO layer in the proposed method. This was because the proposed method was meant to detect only small objects; hence, a multi-scale prediction was not required. A more detailed schematic of the architecture is presented in Fig. 2.

Fig. 2. The proposed method's architecture

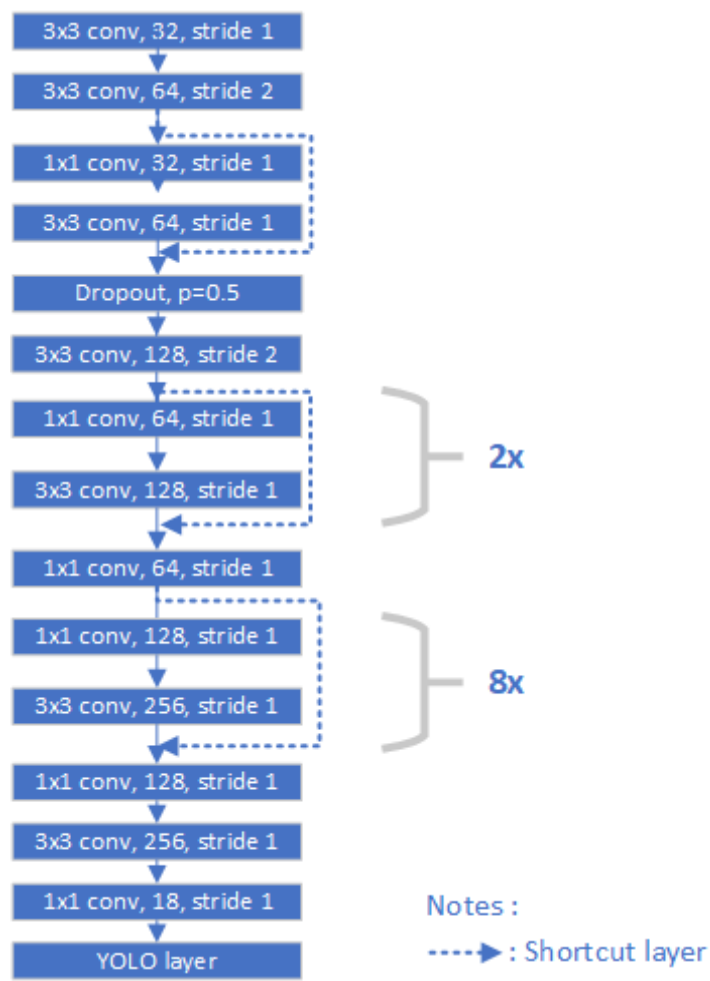

The reason why the number of layers used in the proposed architecture was less than that of YOLOv3 was to increase the training and testing speeds. The maximum filter size was 3 × 3 because the objects were very close to one another. This small filter size improved the speed while maintaining its accuracy. On the other hand, to increase the accuracy, the network resolution was increased to 640 × 640. It was higher than the resolution of the original YOLOv3 architecture [15] and much higher than the widespread YOLOv3 Bochkovskiy implementation [16].

We put a dropout layer just after the first shortcut layer. The threshold was set to half (0.5), which means any node with weight less than half is dropped. This layer was crucial to prevent overfitting and to increase speed while maintaining accuracy. Our code can be accessed at `https://github.com/pHidayatullah/DeepSperm`.

## 2.4 Parameters

Besides neural network architecture, the network parameters were also critical to obtain a faster training speed and a higher accuracy. We set the batch size to 64 and subdivisions to 16. To prevent overfitting, in addition to the dropout layer, the momentum parameter was used to penalize a substantial weight change from one iteration to another, whereas the decay parameter used for penalizing enormous weights. We set the momentum to 0.9 and the decay parameter to 0.0005.

For the same purpose, we fine-tuned the learning rate. In this study, we set the default learning rate to 0.001, with burn-in (warming up) until 250 iterations, which was 6.5% of the total iterations: 4000. We also set the learning rate decay, with a factor of 0.1, after 1000 iterations, which is 25% of the total number of iterations.

Owing to the relatively small dataset, we generated more data from the existing data using data augmentation. The parameters to specify how the data augmentation works were saturation, exposure, and hue. We set saturation to 1.5, exposure to 1.5, and hue to 0.1.

## 2.5 Training and testing environment

We used two different system environments for training and testing. Both systems used Ubuntu 16.04 LTS operating system. Table 2 shows the specification comparison.

Table 2. Hardware specification comparison

|  | Training | Testing |
|---|---|---|
| Model | Server | PC |
| Processor | Intel(R) Xeon(R) Gold 6136 @ 3.00GHz | Intel Core i7 8700 @3.2 GHz |
| RAM | 385 GB | 16 GB |
| GPU | NVIDIA Titan V 12GB GPU RAM | NVIDIA GeForce RTX 2070 8GB GPU RAM |

## 2.6 Training

Following Tajbakhsh et al. [17] claimed that using a pre-trained model consistently outperformed training from scratch, we used a pre-trained model called darknet53.conv.74. It contained convolutional weights trained on ImageNet available in the YOLOv3 repository [15]. For comparison, we trained the proposed architecture on the original darknet framework [15] and Bochkovskiy darknet implementation [16]. For Mask R-CNN, we used the popular mask R-CNN Matterport implementation [18], which was trained with its original parameters.

Bochkovskiy [16] recommended using the number of iterations as many as 2000 times the number of classes. Because one class was used, the recommended number of iterations was 2000. However, we trained the network 4000 times (twice the recommendation) just in case we found the best weights after 2000 iterations.

## 2.7 Inference/testing

We performed the testing of all the methods on the validation dataset as well as on the test dataset. We used mAP@50 as the metric of accuracy so that we can directly compare the proposed method accuracy with those of others. We used [16] for calculating the mAP; we then recorded the results for analysis.

In the testing phase, we tested the proposed method on the original darknet framework [15] as well as on the Bochkovskiy darknet implementation [16]. On the Bochkovskiy implementation, we turned on the CUDNN_HALF option. This option allowed for the use of Tensor Cores of the GPU card to speed up the testing process.

## 3 Results and Discussion

### 3.1 Accuracy

The proposed method achieved 93.77 mAP on the validation dataset and 86.91 mAP on the test dataset, which was higher than that of YOLOv3 by 3.86 mAP and 16.66 mAP, respectively. Fig. 3, 4, and 5 present the comparison of the results.

The digital image processing approach performance was not as good as deep learning-based approaches. Mask R-CNN was mentioned as the most accurate [10], [19]. However, it was struggling to achieve higher accuracy. This was because of the relatively small-size training dataset. Therefore, it overfitted as the training process run.

YOLOv3 Bochkovskiy implementation [23] used a $416 \times 416$ network resolution, whereas the original YOLOv3 used a $608 \times 608$ resolution. The original YOLOv3 achieved 89.91 mAP on the validation dataset. However, the accuracy dropped on the test dataset. To increase the accuracy, the resolution was increased to $640 \times 640$. With this resolution, the input video frames were divided into a $640 \times 640$ grid, which reduced the grid size. It was the main key to our proposed method accuracy.

Fig. 3. Results comparison on one video frame from the validation dataset

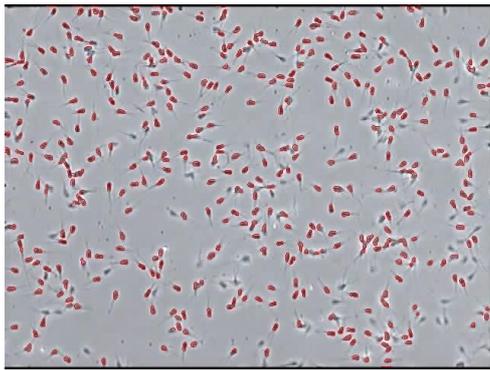
a. Digital image processing (74.72 mAP)

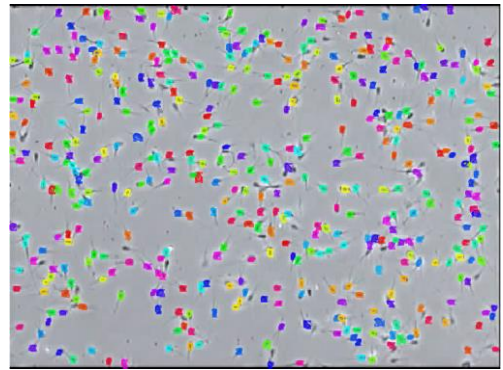
b. Mask R-CNN (84.57 mAP)

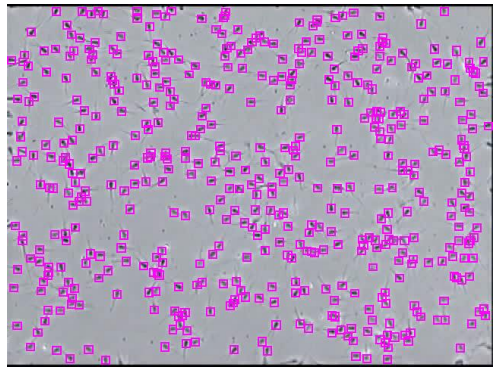
c. YOLOv3 (89.70 mAP)

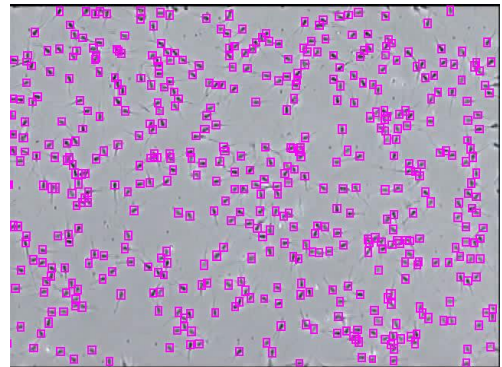
d. DeepSperm (93.74 mAP)

Fig. 4. Results comparison on one video frame from the test dataset

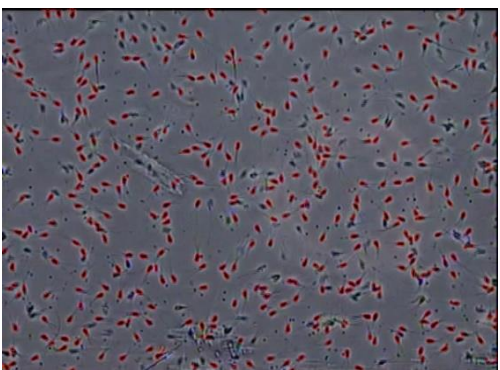
a. Digital image processing (55.98 mAP)

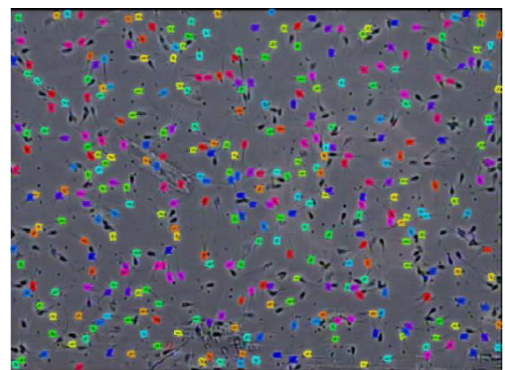
b. Mask R-CNN (64.77 mAP)

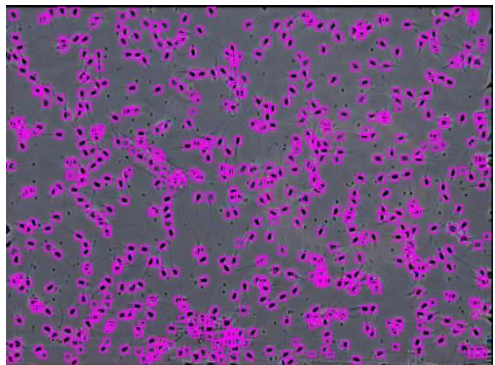
c. YOLOv3 (67.78 mAP)

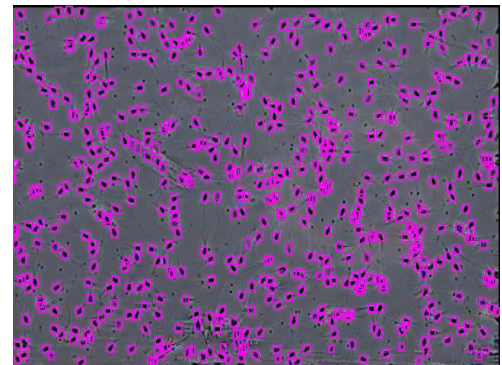
d. DeepSperm (84.54 mAP)

Fig. 5. Accuracy comparison of all the methods

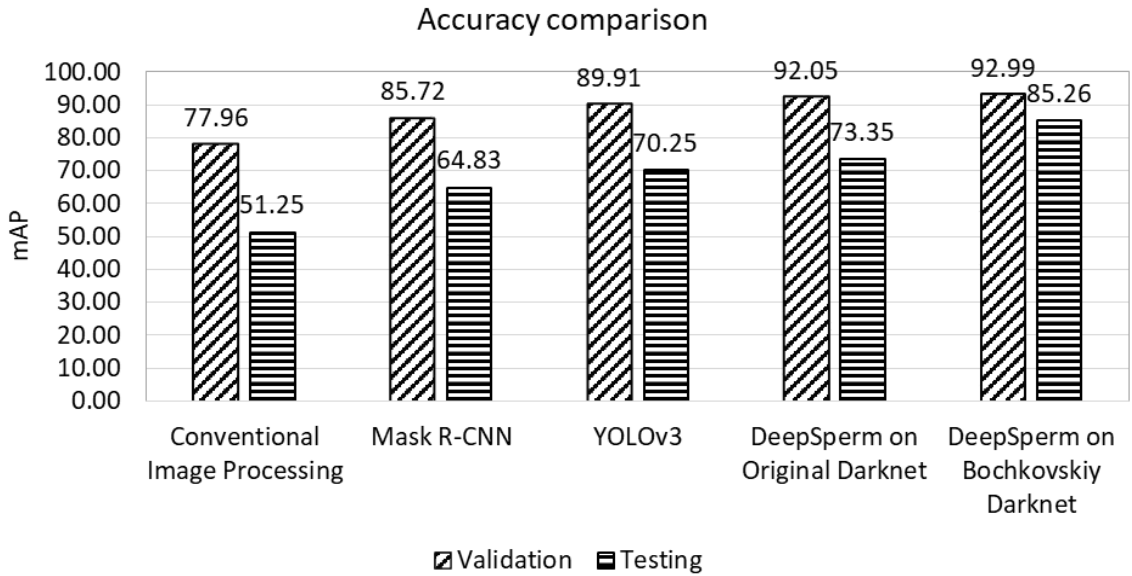

## 3.2 Partial occlusion handling

Partial occlusion is one of the main challenges that causes limited detection accuracy. However, compared to other methods, our proposed method had been excellent at handling partial occlusion. Fig. 6 shows the video frame, and Table 3 presents the comparison details.

In the first case, YOLOv3 and Mask R-CNN failed to handle the partial occlusion of the top sperm cells. Our proposed method and the conventional image processing-based method were able to detect the partially occluded sperms correctly. However, the conventional image processing-based method could not detect some noticeable sperm cells.

In the second case, our proposed method was able to detect the sperms accurately. YOLOv3 was able to detect the occluded sperm cells at the bottom. However, it produced one false positive. The other methods still suffered.

Fig. 6. The partial occlusion analysis

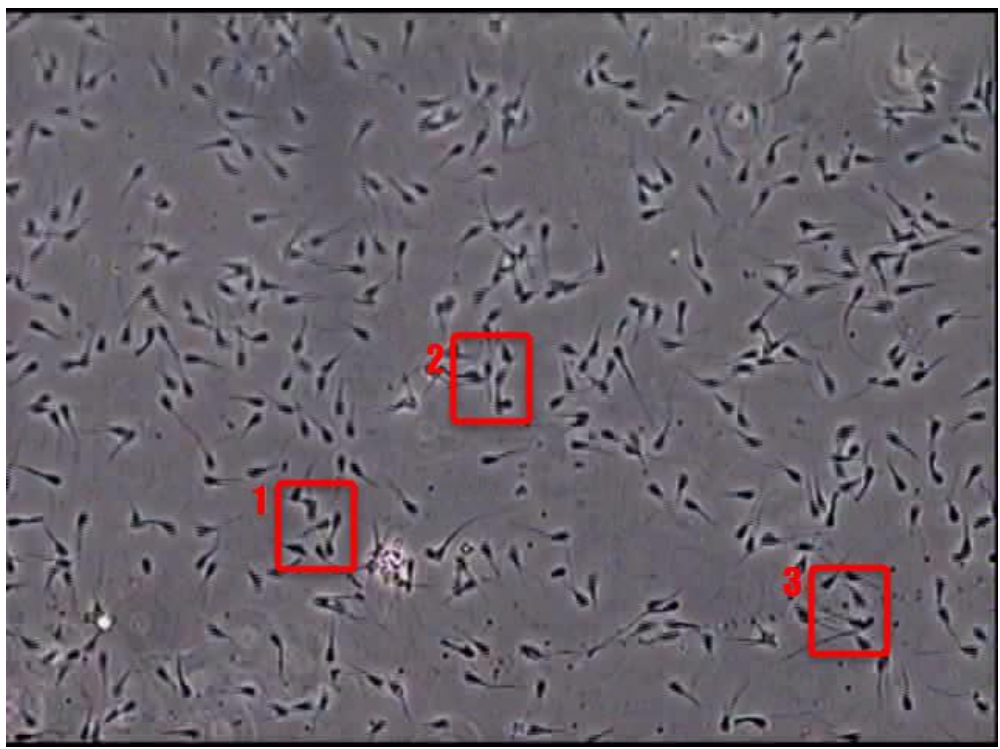

Table 3. Partial occlusion handling comparison

| Case No | Ground truth | Conventional Image Processing | Mask R-CNN | YOLOv3 | DeepSperm |
|---|---|---|---|---|---|
| 1 | 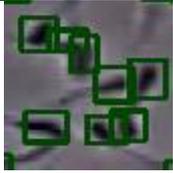 | 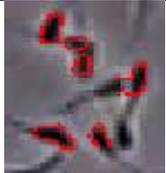 | 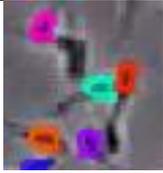 | 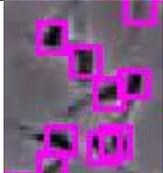 | 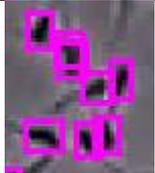 |
| 2 | 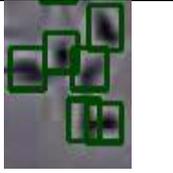 | 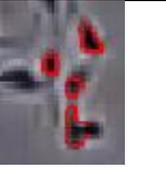 | 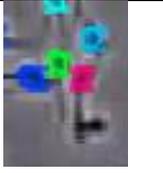 | 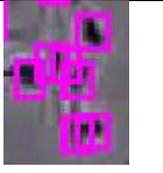 | 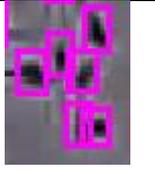 |
| 3 | 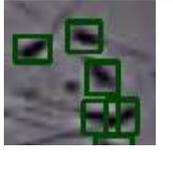 | 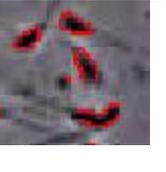 | 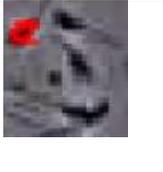 | 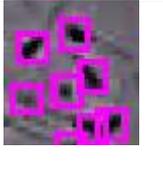 | 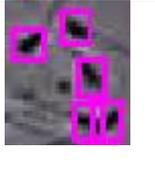 |

In the third case, YOLOv3 detected some small artifacts as sperm cells. Mask R-CNN suffered the most. However, our proposed method was able to detect them accurately.

## 3.3 Artifacts handling

Artifacts are also the main challenges that lead to limited detection accuracy. For example, in Fig. 7, there were tiny marks. They had the same color as the sperm cells but smaller in size.

In YOLO9000 [20], the authors increased the recall of YOLO [13] by using anchor boxes. However, they obtained a small decrease in accuracy (mAP) [20]. That was the reason why YOLO based detector produces a significant amount of false positives. Table 4 indicates that YOLOv3 regarded seven artifacts as sperm cells.

Fig. 7. Artifacts on a sample

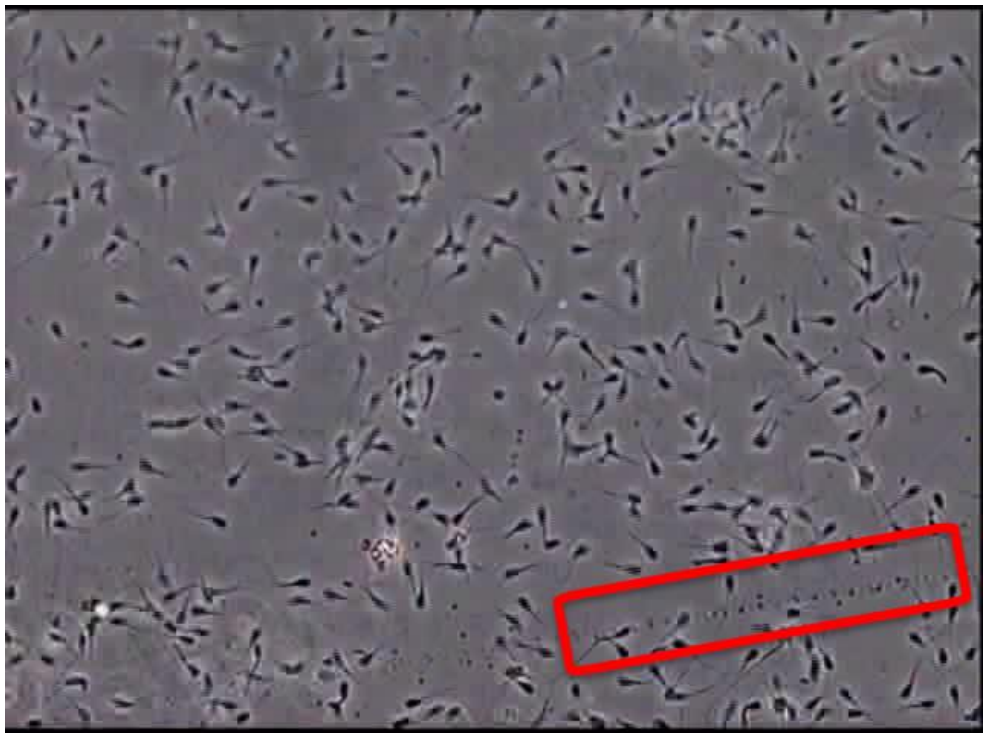

Table 4. Artifacts handling comparison

| Method | Result |
| --- | --- |
| Ground truth | 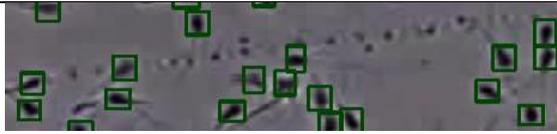 |
| Conventional image processing | 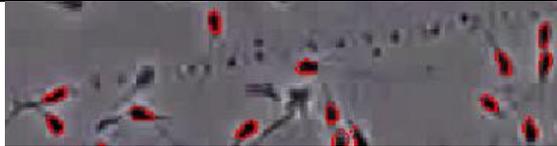 |
| Mask R-CNN | 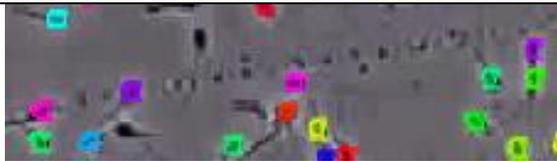 |
| YOLOv3 | 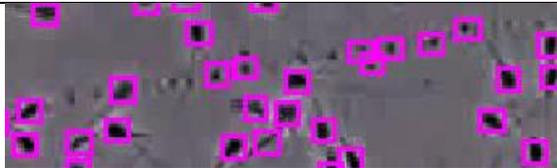 |
| DeepSperm | 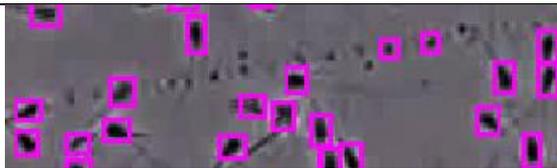 |

We have managed to reduce false positives by increasing network resolution. Compared to YOLOv3, we can reduce the number of false positives so that only two of these artifacts were regarded as sperm cells. Mask R-CNN and the image processing based detection method were better in handling these artifacts. However, both were struggling to detect some noticeable sperm cells.

### 3.4 Overfitting handling

The case in this study was vulnerable to overfitting because the training dataset was relatively small. Mask R-CNN detection on the test dataset dropped by 20.89 mAP points compared to detection on the validation dataset. The detection performed by YOLOv3 dropped by 19.66 mAP points.

YOLOv3 used batch normalization in every convolutional layer as well as in ours. Batch normalization was considered sufficient without any form of other regularizations [20]. Therefore, the dropout layers were removed since YOLO9000 [20]. However, we observed that using only batch normalization was not sufficient for reducing overfitting. It was because of the small number of samples in the training dataset. As a solution, we utilized the dropout layer with a threshold of 0.5. There were two recommended positions of the dropout layer: the first at every layer and the second at the first layer only [21]. We observed that putting the dropout layer at the first shortcut layer yielded much better results than putting it at every shortcut layer.

In addition, based on the dataset histogram analysis, we performed data augmentation to increase the number of training data with varying the data according to its saturation, exposure, and hue. The result indicated that the detection accuracy on the test dataset, as well as on the validation dataset, could be increased. With only a 6.86 mAP gap, we achieved up to 93.77 mAP and 86.91 mAP on the validation and test datasets, respectively.

### 3.5 Speed

Speed is another highly critical criterion for object detection. Mask R-CNN achieved only 0.19 fps (the slowest) even though it used a GPU. This was because of the two-stages approach and the large size of the network. The conventional image processing method achieved 1.54 fps (8.1 times faster than Mask R-CNN), though it only had the CPU mode. However, it delivered a lower accuracy. YOLOv3 significantly increased the speed (10 times that of the conventional image processing method). In addition to its speed, the accuracy was also increased. Fig. 8 presents the speed comparison.

Fig. 8. Speed comparison of all the methods during the test phase

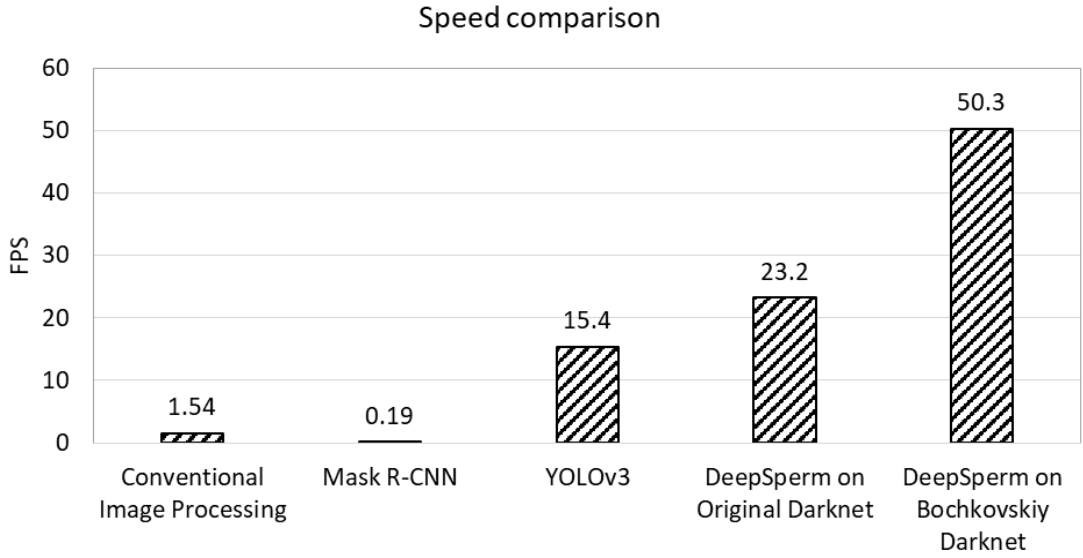

To increase the speed, we used a smaller network with a higher network resolution, which made the speed 1.51 times faster than the speed of YOLOv3. To further increase the speed, we used the Bochkovskiy darknet implementation [16], which employed Tensor Cores, and speed 3.26 times that of YOLOv3 was achieved.

In the training phase, Mask R-CNN needed 155.9 s for an epoch. YOLOv3 was 19.7 times faster than Mask R-CNN in training, which was a significant boost. We achieved a higher speed: 1.21–1.4 times faster than the YOLOv3.

### 3.6  Failure case analysis

In general, our proposed method has been able to detect sperm cells better than the other methods, summarized in Tables 3 and 4. However, we still encountered some detection errors. Table 5 presents some of the detection failures which were highlighted by the arrows.

Table 5. Some detection failures

| Case | Original frame | Detection result |
|---|---|---|
| 1. Failure example from the validation dataset (video 1) | 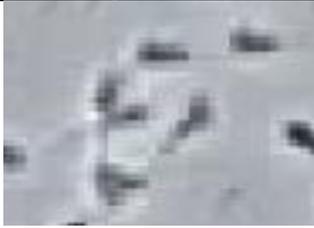 | 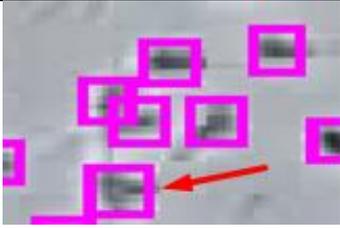 |
| 2. Failure example from the test dataset (occlusion and artifacts) | 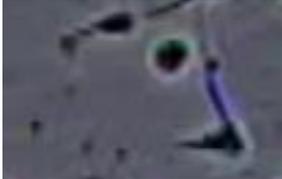 | 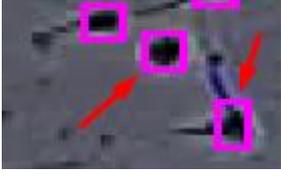 |
| 3. Failure example from the test dataset (false positive) | 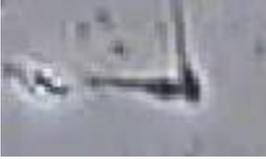 | 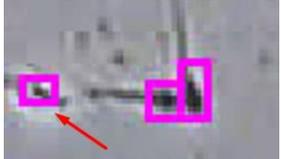 |
| 4. Failure example from the test dataset (false negative) | 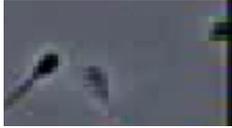 | 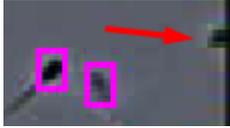 |

To summarize, these were the reasons of the failures: the sperm were almost fully occluded (first and second cases), the artifacts were more or less similar to a sperm cell (second case and third cases), the sperm head was attached to the video frame border which made it appeared as a dark spot/artifact (fourth case).

### 3.7 Overall comparison

For clarity, we compared the performances of all the methods in a single table (Table 6). It was inferred that Mask R-CNN [10] delivered a better accuracy than the previous YOLO-based model [13], [19], [20]. However, it was also claimed that YOLOv3 (the newest version of YOLO) delivered a better accuracy in detecting small objects, compared to other methods [11]. Our experiments also confirmed what YOLOv3 claimed. Therefore, because sperm-cell detection is a small-object detection, we add a column to the table containing the performance improvement of our proposed method to YOLOv3.

Table 6. Overall results comparison

| Criteria | Conventional Image Processing | Mask R-CNN | YOLOv3 | DeepSperm | | |
|---|---|---|---|---|---|---|
| | | | Original Darknet | Original Darknet | Bochkovskiy Darknet | Improvement to YOLOv3 |
| Number of layers | n/a | 235 | 106 | **42** | **42** | 2.52x smaller |
| Weights file size (MB) | n/a | 255.9 | 240.5 | **14.2** | **14.2** | 16.94x smaller |
| Best weights | n/a | 3971 | 3600 | **900** | 1200 | 3 - 4x faster to converge |
| Training time/epoch (s) | n/a | 155.9 | 7.9 | 6.51 | **5.64** | 1.21 - 1.4x faster |
| GPU RAM need (GB) | n/a | 11.7 | 7.4 | **5.7** | **5.7** | 1.3x lesser |
| Validation set accuracy (mAP@50) | 77.96 | 85.72 | 89.91 | 92.26 | **93.77** | 2.35 – 3.86 |
| Test set accuracy (mAP@50) | 51.25 | 64.83 | 70.25 | 82.23 | **86.91** | 11.98 – 16.66 |
| Fps | 1.54 | 0.19 | 15.4 | 23.2 | **50.3** | 1.51- 3.26x faster |
| Testing time/image (ms) | 649.35 | 5,222.04 | 46.91 | 19.78 | **14.89** | 2.37 - 3.15x faster |

## 4 Conclusions

This study proposed a deep CNN architecture, with its hyper-parameters and configurations detailed in the material and methods' section, for robust detection of bull sperm cells. It was robust to partial occlusion, artifacts, vast number of moving objects, object's tiny size, low contrast, blurred objects, and different lighting conditions.

To summarize, the proposed method surpassed all the methods in terms of accuracy and speed. The proposed method achieved 86.91 mAP on the test dataset, 16.66 mAP points higher than the state-of-the-art method (YOLOv3). In terms of speed, the proposed method achieved real-time performance with up to 50.3 fps, which was 3.26 times faster than the state-of-the-art method. Our training time was also faster, up to 1.4 times that of the state-of-the-art method. With that performance, it eventually needed a small training set containing only 40 video frames.

The proposed architecture was also much smaller than YOLOv3 as well as Mask R-CNN. There were some advantages to such architectures. The training and testing times were fast, less GPU memory was needed (1.3 times lesser than YOLOv3 and 2.05 times lesser than Mask R-CNN), and less amount of file storage was needed (weights file's size was 16.94 times smaller than the YOLOv3 weights file and 18.02 times smaller than the Mask R-CNN weights file).

In the future, we want to apply the proposed method for human sperm cell detection. We also want to test it for different cases such as detecting blood cells, bacteria, or any other biomedical case. We believe that this architecture shall perform well in these cases too.

## 5  Acknowledgments

This research was funded by Lembaga Pengelola Dana Pendidikan (LPDP) RI [grant number PRJ-6897 /LPDP.3/2016]. The authors would like to thank ITB World Class University (WCU) program, Prof. Ryosuke Furuta, Mr. Ridwan Ilyas, and Mr. Kurniawan Nur Ramdhani.